\def\year{2021}\relax
\documentclass[letterpaper]{article} 
\usepackage{aaai21}  
\usepackage{times}  
\usepackage{helvet} 
\usepackage{courier}  
\usepackage[hyphens]{url}  
\usepackage{graphicx} 
\usepackage{natbib}  
\usepackage{caption} 
\usepackage{xspace}
\usepackage[utf8]{inputenc}
\usepackage{enumitem}
\usepackage{xcolor}
\usepackage{stmaryrd}
\usepackage{comment}

\usepackage{adjustbox}         
\usepackage{booktabs,multirow,siunitx}  
\usepackage{mathptmx}          

\usepackage{xspace}
\usepackage{empheq}

\usepackage{amsfonts,amsmath,amssymb,amsthm,bm,latexsym}
\usepackage{verbatim,color,epsfig,graphics,url}
\usepackage[noend,linesnumberedhidden,ruled,resetcount,noline]{algorithm2e}
\usepackage{algorithmic}
\usepackage{dashbox}
\usepackage{fancybox}
\usepackage{subcaption}

\usepackage{pgf}
\usepackage{tikz}
\usepackage{forest}

\usepackage{soul}

\graphicspath{{plots/}}
\DeclareGraphicsExtensions{.pdf}

\usetikzlibrary{arrows,decorations.pathmorphing,decorations.footprints,fadings,calc,trees,mindmap,shadows,decorations.text,patterns,positioning,shapes,matrix,fit}
\usetikzlibrary{arrows,shadows,backgrounds}
\usetikzlibrary{arrows.meta}
\usetikzlibrary{positioning,fit}
\usetikzlibrary{automata}
\usetikzlibrary{matrix}
\usetikzlibrary{shapes.symbols,shapes.misc,shapes.arrows}
\usetikzlibrary{matrix,chains,scopes,decorations.pathmorphing}

\DeclareMathAlphabet{\mathcal}{OMS}{cmsy}{m}{n}

\newcommand{\papertitle}{A Scalable Two Stage Approach to Computing Optimal Decision Sets}

\title{\papertitle%
  \thanks{This work is supported in part by the AI Interdisciplinary
    Institute ANITI, funded by the French program ``Investing for the
    Future - PIA3'' under Grant agreement no.~ANR-19-PI3A-0004.}%
}
\author{
  Alexey Ignatiev\textsuperscript{\rm 1},
  Edward Lam\textsuperscript{\rm 1,2},
  Peter J. Stuckey\textsuperscript{\rm 1},
  Joao Marques-Silva\textsuperscript{\rm 3} \\
}%

\affiliations{%
  \textsuperscript{\rm 1}Monash University, Melbourne, Australia \\
  \textsuperscript{\rm 2}CSIRO Data61, Melbourne, Australia \\
  \textsuperscript{\rm 3}ANITI, IRIT, CNRS, Toulouse, France \\
  \{alexey.ignatiev,edward.lam,peter.stuckey\}@monash.edu,
  joao.marques-silva@irit.fr
}

\definecolor{tyellow1}{HTML}{FCE94F}
\definecolor{tyellow2}{HTML}{EDD400}
\definecolor{tyellow3}{HTML}{C4A000}
\definecolor{torange1}{HTML}{FCAF3E}
\definecolor{torange2}{HTML}{F57900}
\definecolor{torange3}{HTML}{C35C00}
\definecolor{tbrown1}{HTML}{E9B96E}
\definecolor{tbrown2}{HTML}{C17D11}
\definecolor{tbrown3}{HTML}{8F5902}
\definecolor{tgreen1}{HTML}{8AE234}
\definecolor{tgreen2}{HTML}{73D216}
\definecolor{tgreen3}{HTML}{4E9A06}
\definecolor{tblue1}{HTML}{729FCF}
\definecolor{tblue2}{HTML}{3465A4}
\definecolor{tblue3}{HTML}{204A87}
\definecolor{tpurple1}{HTML}{AD7FA8}
\definecolor{tpurple2}{HTML}{75507B}
\definecolor{tpurple3}{HTML}{5C3566}
\definecolor{tred1}{HTML}{EF2929}
\definecolor{tred2}{HTML}{CC0000}
\definecolor{tred3}{HTML}{A40000}
\definecolor{tlgray1}{HTML}{EEEEEC}
\definecolor{tlgray2}{HTML}{D3D7CF}
\definecolor{tlgray3}{HTML}{BABDB6}
\definecolor{tdgray1}{HTML}{888A85}
\definecolor{tdgray2}{HTML}{555753}
\definecolor{tdgray3}{HTML}{2E3436}

\newtheorem{theorem}{Theorem}

\theoremstyle{definition}
\newtheorem{definition}[theorem]{Definition}
\theoremstyle{remark}
\newtheorem{example}{Example}

\definecolor{gray}{rgb}{.4,.4,.4}
\definecolor{midgrey}{rgb}{0.5,0.5,0.5}
\definecolor{middarkgrey}{rgb}{0.35,0.35,0.35}
\definecolor{darkgrey}{rgb}{0.3,0.3,0.3}
\definecolor{darkred}{rgb}{0.7,0.1,0.1}
\definecolor{midblue}{rgb}{0.2,0.2,0.7}
\definecolor{darkblue}{rgb}{0.1,0.1,0.5}
\definecolor{darkgreen}{rgb}{0.1,0.5,0.1}
\definecolor{defseagreen}{cmyk}{0.69,0,0.50,0}

\newcommand{\fml}[1]{{\mathcal{#1}}}

\newcommand{\mbf}[1]{\ensuremath\mathbf{#1}}
\newcommand{\mbb}[1]{\ensuremath\mathbb{#1}}

\newcommand{\dcare}{\ensuremath{\mathfrak{u}}\xspace}

\newcommand{\pds}{\ensuremath{{\phi}_\oplus}\xspace}
\newcommand{\nds}{\ensuremath{{\phi}_\ominus}\xspace}

\newcommand{\mpr}[2]{\ensuremath{ruler_{{#1}}^{{#2}}}\xspace}
\newcommand{\mprb}[2]{\ensuremath{ruler_{{#1}}^{{#2}}\text{\it +}b}\xspace}
\newcommand{\epos}{\ensuremath{\fml{E}_\oplus}\xspace}
\newcommand{\eneg}{\ensuremath{\fml{E}_\ominus}\xspace}
\newcommand{\tpos}{\ensuremath{\fml{T}_\oplus}\xspace}
\newcommand{\tneg}{\ensuremath{\fml{T}_\ominus}\xspace}

\tikzset{
  0 my edge/.style={densely dashed, my edge},
  my edge/.style={-{Stealth[]}},
}

\newcommand{\ignore}[1]{}

\newcommand{\frmeq}[1]{\begin{empheq}[box={\fboxsep=1.5pt\doublebox}]{flalign*}#1\end{empheq}}

\begin{document}

\maketitle

\begin{abstract}
  Machine learning (ML) is ubiquitous in modern life.
  Since it is being deployed in technologies that affect our privacy
  and safety, it is often crucial to understand the reasoning behind
  its decisions, warranting the need for \emph{explainable AI}.
  Rule-based models, such as decision trees, decision lists, and
  \emph{decision sets}, are conventionally deemed to be the most
  interpretable.
  Recent work uses propositional satisfiability (SAT) solving (and its
  optimization variants) to generate minimum-size decision sets.
  Motivated by limited practical scalability of these earlier methods,
  this paper proposes a novel approach to learn minimum-size decision
  sets by enumerating individual rules of the target decision set
  independently of each other, and then solving a set cover problem to
  select a subset of rules.
  The approach makes use of modern maximum satisfiability and integer
  linear programming technologies.
  Experiments on a wide range of publicly available datasets
  demonstrate the advantage of the new approach over the state of the
  art in SAT-based decision set learning.
\end{abstract}

\section{Introduction} \label{sec:intro}

Rapid advances in artificial intelligence and, in
particular, in machine learning (ML), have influenced all aspects of human lives.
Given the practical achievements and the overall success of modern
approaches to
ML~\cite{bengio-nature15,jordan-science15,silver-nature15,taward18},
one can argue that it will prevail as a generic
computing paradigm and it will find an ever growing range of practical
applications.
Unfortunately, the most widely used ML models are opaque, which makes it
hard for a human decision-maker to comprehend the outcomes of such
models.
This motivated efforts on validating the operation of ML
models~\cite{kwiatkowska-ijcai18,barrett-cav17} but also on devising
approaches to \emph{explainable artificial intelligence}
(XAI)~\cite{guestrin-kdd16,lundberg-nips17,monroe-cacm18}.

One of the major lines of work in XAI is devoted to training
logic-based models, e.g.\ decision
trees~\cite{bessiere-cp09,nipms-ijcai18,rudin-nips19,nijssen-aaai20},
decision lists~\cite{rudin-kdd17a,rudin-mpc18} or decision
sets~\cite{leskovec-kdd16,ipnms-ijcar18,meel-cp18,meel-aies19,yisb-cp20},
where concise explanations can be obtained directly from the model.
This paper focuses on the decision set (DS) model, which comprises an
unordered set of \emph{if-then} rules.

One of the advantages of decision sets over other rule-based models is
that rule independence makes it straightforward to explain a
prediction: a user can pick any rule that ``fires'' the prediction and
the rule itself serves as the explanation.
As a result, generation of minimum-size decision sets is of great
interest.
Recent work proposed SAT-based methods for generating minimum-size
decision sets by solving a sequence of problems that determine whether
a decision set of size $K$ exists, with $K$ being the number of
rules~\cite{ipnms-ijcar18} or the number of
literals~\cite{yisb-cp20} s.t.\ the decision set agrees with the
training data.
Unfortunately, scalability of both approaches is limited due to the
large size of the propositional encoding.

Motivated by this limitation, our work proposes a novel approach that
splits the DS generation problem into two parts: (1)~exhaustive rule
enumeration and (2)~computing a subset of rules agreeing with the
training data.
In general, this novel approach enables a significantly more compact
propositional encoding, which makes it scalable for problem instances
that are out of reach for the state of the art.
The proposed approach is inspired by the standard setup used in
two-level logic
minimization~\cite{quine-amm52,quine-amm55,mccluskey-bstj56}.
While the first part can be done using enumeration of maximum
satisfiability (MaxSAT) solutions, the second part is reduced to the
set cover problem, for which integer linear programming (ILP) is
effective.
Experiments on a wide range of datasets indicate that this approach
outperforms the state of the art.  The remainder of this paper
presents these developments in detail.
%

\section{Preliminaries} \label{sec:prelim}


\paragraph{Satisfiability and Maximum Satisfiability.}
%
We use the standard definitions for propositional satisfiability (SAT)
and maximum satisfiability (MaxSAT) solving~\cite{sat-handbook09}.
SAT and MaxSAT admit Boolean variables.
A \emph{literal} over Boolean variable $x$ is either the variable $x$
itself or its negation $\neg{x}$.
(Given a constant parameter $\sigma\in\{0,1\}$, we use notation
$x^\sigma$ to represent literal $x$ if $\sigma=1$, and to represent
literal $\neg{x}$ if $\sigma=0$.)
A \emph{clause} is a disjunction of literals. A \emph{term} is a
conjunction of literals.
A propositional formula is said to be in \emph{conjunctive normal
form} (CNF) or \emph{disjunctive normal form} (DNF) if it is a
conjunction of clauses or disjunction of terms, respectively.
Whenever convenient, clauses and terms are treated as sets of
literals.
%
Formulas written as sets of sets of literals (either in CNF or DNF)
are described as \emph{clausal}.

%

We will make use of partial maximum
satisfiability (Partial MaxSAT)~\cite[Chapter~19]{sat-handbook09},
which can be formulated as follows.
A partial CNF formula can be seen as a conjunction of \emph{hard}
clauses $\fml{H}$ (which must be satisfied) and \emph{soft} clauses
$\fml{S}$ (which represent a preference to satisfy those clauses).
%
%
%
The Partial MaxSAT problem consists in finding an assignment that
satisfies all the hard clauses and maximizes the total number of
satisfied soft clauses.

\paragraph{Classification Problems and Decision Sets.}
We follow the notation used in earlier
work~\cite{bessiere-cp09,leskovec-kdd16,ipnms-ijcar18,yisb-cp20}.
Consider a set of features $\fml{F}=\{1,\ldots,K\}$.
The domain of possible values for feature $r \in [K]$ is $D_r$.
The complete space of feature values (or \emph{feature
space}~\cite{han-bk12}) is $\mbb{F}\triangleq\prod_{r=1}^{K} D_r$.
The vector $\mbf{f}=(f_1,\ldots,f_K)$ of $K$ variables $f_r \in D_r$
refers to a point in $\mbb{F}$.
Concrete (constant) points in $\mbb{F}$ are denoted by
$\mbf{v}=(v_1,\ldots,v_K)$, with $v_r\in D_r$.
%
For simplicity, all the features are assumed to be binary, i.e.\
$D_r=\{0,1\}, \forall r \in [K]$; categorical and ordinal features can
be mapped to binary features using standard
techniques~\cite{scikitlearn-full}.
Therefore, whenever convenient, a Boolean \emph{literal} on a feature
$r$ can be represented as $f_r$ (or $\neg{f_r}$, resp.), denoting that
feature $f_r$ takes value 1 (value 0, resp.).

Consider a standard classification scenario with training data
$\fml{E}=\{e_1,\ldots,e_M\}$.
A data instance (or \emph{example}) $e_i\in\fml{E}$ is a pair
$(\mbf{v}_i,c_i)$ where $\mbf{v}_i\in\mbb{F}$ is a vector of feature
values and $c_i\in\fml{C}$ is a class.
An example $e_i$ can be seen as \emph{associating} a vector of feature
values $\mbf{v}_i$ with a class $c_i\in\fml{C}$.
%
%
%
This work focuses on binary classification problems, i.e.
$\fml{C}=\{\ominus,\oplus\}$ but the proposed ideas are easily
extendable to the case of multiple classes.
%
%
Given example $e_i=(\mbf{v}_i,c_i)\in\fml{E}$ and the $r^\text{th}$
component $v_{ir}$ of $\mbf{v}_i$, literal $f_r^{1-v_{ir}}$ is said to
\emph{discriminate} $e_i$ because
$f_r^{1-v_{ir}}\triangleq\neg{v_{ir}}$, i.e. $f_r^{1-v_{ir}}$
falsifies example $e_i$ when it is considered as a conjunction of
feature literals.
The concept of example discrimination can be extended to terms.

Examples $e_i,e_j \in \fml{E}$ associating the same set of feature
values with the opposite classes are referred to as
\emph{overlapping}.
We assume wlog.  that the training data $\fml{E}$ is \emph{perfectly
classifiable}, i.e.\ $\fml{E}$ partially defines a Boolean function
$\phi:\mbb{F}\rightarrow\fml{C}$ --- in other words, there are no
overlapping examples in $\fml{E}$.
Otherwise, either $e_i$ or $e_j$ can be removed from dataset
$\fml{E}$, incurring an error of 1.
In general, repeated \emph{``collisions''} can be resolved by taking
the majority vote, which results in highest possible accuracy on
training data.


%
\ignore{
%
  -- in other words, $(\forall r\in[K]\
  \mbf{v}_i[r]=\mbf{v}_j[r])\Rightarrow c_i=c_j$ (alternatively and in
  order to deal with \emph{inconsistencies}, one can assume $\phi$ to
  be a relation defined on subset of $\mbb{F}\times\fml{C}$).
}

The objective of classification in ML is to devise a function
$\hat{\phi}$ that matches the actual function $\phi$ on the training
data $\fml{E}$ and generalizes \emph{suitably well} on unseen test
data~\cite{furnkranz-bk12,han-bk12,mitchell-bk97,quinlan-bk93}.
In many settings, function $\hat{\phi}$ is not required to match
$\phi$ on the complete set of examples $\fml{E}$ and instead an
\emph{accuracy} measure is considered.
%
%
Furthermore, in classification problems one conventionally has to
optimize with respect to (1) the complexity of $\hat{\phi}$, (2) the
accuracy of the learnt function (to make it match the actual function
$\phi$ on a maximum number of examples), or (3) both.
As this paper assumes that the training data does not have overlapping
instances, we aim solely at minimizing the representation size of the
target ML models.

\ignore{
  work has considered various representations of the target function
  $\hat{\phi}$.
  Some of the representations are known to be $\ldots$
  \cite{citations}.
}
This paper focuses on learning representations of $\hat{\phi}$
corresponding to \emph{decision sets}
(DS)~\cite{leskovec-kdd16,ipnms-ijcar18,meel-cp18,meel-aies19,yisb-cp20}.
A decision set is an \emph{unordered set} of \emph{rules}.
Each rule $\pi$ is from the set $\fml{R}=\prod_{r=1}^{K}\{f_r,\neg
f_r,\dcare\}$, where \dcare represents a \emph{don't care} value.
For each example $e\in\fml{E}$, a rule of the form $\pi \Rightarrow
c$, $\pi \in \fml{R}$, $c \in \fml{C}$ is interpreted as ``if the
feature values of example $e$ agree with $\pi$ then the rule predicts
that example $e$ has class $c$''.
Hereinafter, we will be dealing with learning minimum-size decision
sets, with the size measure being either the number of rules in the
decision set or the total number of literals in it (sometimes referred
to as \emph{total size}).
Note that because rules in decision sets are unordered, some rules may
\emph{overlap}, i.e.\ multiple rules $\pi_i \in \fml{R}$ may agree
with some instance of the feature space $\mbb{F}$.
It may also happen that \emph{none of the rules} of a decision set
apply to some instances of $\mbb{F}$.
\begin{example}\label{ex:data}
  Consider the following dataset of four data instances representing
  the \emph{``to date or not to date?''} example
  by~\citeauthor{domingos-mabook15}~(\citeyear{domingos-mabook15}):

\newcommand{\blue}{\color{tblue3}}
\newcommand{\red}{\color{tred3}}
\newcommand{\green}{\color{tgreen3}}

\begin{center}
\small
\begin{tabular}{cccccc}
\toprule
\textbf{\#} & \textbf{Day} &
\textbf{Venue} & \textbf{Weather} &
\textbf{TV Show} & \textbf{\em Date?} \\
\midrule
$e_1$ & Weekday & Dinner & Warm & Bad & No \\
$e_2$ & Weekend & Club & Warm & Bad & Yes \\
$e_3$ & Weekend & Club & Warm & Bad & Yes \\
$e_4$ & Weekend & Club & Cold & Good & No \\
\bottomrule
\end{tabular}
\end{center}
This data serves to predict whether a friend accepts an
invitation to go out for a date given various circumstances.
An example of a valid decision set for this data is the following:

\frmeq{
  \small
  \begin{array}{ll}
    \textbf{IF } \blue \text{ TV Show}=\text{Good} & \textbf{THEN }
    \red \text{ Date}=\text{No} \\
    \textbf{IF } \blue \text{ Day}=\text{Weekday} & \textbf{THEN }
    \red \text{ Date}=\text{No} \\
    \textbf{IF } \blue \text{ TV Show}=\text{Bad } \bm{\land} \text{
    Day}=\text{Weekend} & \textbf{THEN } \green \text{
  Date}=\text{Yes} \\
  \end{array}
}
This DS has 3 rules and a
\emph{total size} of 7 (1 for each literal
on the left and right, or alternatively, 1 for each
literal on the left and 1 for each rule).
It does not exhibit rule overlap for examples
in $\mbb{F}$ while the following decision set does:

\frmeq{
  \small
  \begin{array}{ll}
    \textbf{IF } \blue \text{ TV Show}=\text{Good} & \textbf{THEN }
    \red \text{ Date}=\text{No} \\
    \textbf{IF } \blue \text{ Day}=\text{Weekday} & \textbf{THEN }
    \red \text{ Date}=\text{No} \\
    \textbf{IF } \blue \text{ Weather}=\text{Warm } \bm{\land} \text{
    Day}=\text{Weekend} & \textbf{THEN } \green \text{
  Date}=\text{Yes} \\
  \end{array}
}
Here, the first and third rules overlap for all examples with feature
values $\textit{Weather}=\textit{Warm}$ and $\textit{TV
Show}=\textit{Good}$.
\qed
\end{example}

\section{Related Work} \label{sec:relw}

Rule-based ML models can be traced back to around the 70s and
80s~\cite{michalski-isip69,shwayder-cacm75,rivest-ipl76,breiman-bk84,quinlan-ml86,rivest-ml87}.
To our best knowledge, decision sets first appear as an unordered
variant of decision lists~\cite{rivest-ml87,clark-ml89}
in~\cite{clark-ewsl91}.
The use of logic and optimization for synthesizing a disjunction of
rules matching a given training dataset was first tried
in~\cite{resende-mp92}.
Recently, \cite{leskovec-kdd16} argued that decision sets are more
interpretable than decision trees and decision lists.

Our work builds on~\cite{ipnms-ijcar18,yisb-cp20} where SAT-based
models were proposed for training decision sets of smallest size.
The method of \cite{ipnms-ijcar18} minimized the number of rules in
\emph{perfect} decision sets, i.e.\ those that agree perfectly with
the training data, which is assumed to be consistent;
it was also shown to significantly outperform the smooth local search approach of~\cite{leskovec-kdd16}.
Rule minimization was then followed by minimization of the total
number of literals used in the decision set, which resulted in a
lexicographic approach to the minimization problem.
In contrast, \cite{yisb-cp20} focused on minimizing the total number
of literals in the target DS.
This work showed that minimizing the number of rules is more scalable
for solving the perfect decision set problem since the optimization
measure, i.e.\ the number of rules, is more coarse-grained.
However, minimizing the total number of literals was shown to produce
significantly smaller and, thus, more \emph{interpretable} target
decision sets.
Furthermore, they showed that \emph{sparse} decision sets (minimizing
either the number of literals or the number or rules) provide a user
with yet another way to produce a succinct classifier representation,
by trading off its accuracy for smaller size.
Sparse decision sets were also considered
in~\cite{meel-cp18,meel-aies19} where the authors proposed a MaxSAT
model for representing one target class of the training data.
ILP was also applied to compute a variant of sparse decision
sets~\cite{dash-nips18}.
As was shown in~\cite{yisb-cp20}, sparse decision sets, although are
much easier to compute, achieve lower test accuracy compared to
perfect decision sets.
As a result, the focus of this work is solely on improving scalability
of computing perfect decision sets, i.e.\ sparse models are excluded
from consideration.
%


\section{Decision Sets by Rule Enumeration} \label{sec:rules}
%
Similar to the recent logic-based approaches to learning decision
sets~\cite{ipnms-ijcar18,meel-cp18,meel-aies19}, our approach builds
on state-of-the-art SAT and MaxSAT technology.
These prior works consider a SAT or MaxSAT model that determines
whether there exists a decision set of size $N$ given training data
$\fml{E}$ with $|\fml{E}|=M$ examples.
The problem is solved by iteratively varying size $N$ and making
either a series of SAT calls or one MaxSAT call.
The main limitation of prior work is the encoding formula size, which
is $\fml{O}(N\times M\times K)$, where $N$ is the target size of
decision set (which is determined either as the number of
rules~\cite{ipnms-ijcar18} or as the total number of
literals~\cite{yisb-cp20}), $M$ is the number of training data
instances and $K$ is the number of features in the training data.
This limitation significantly impairs scalability of these approaches
and hence restricts their practical applicability.

In contrast to the aforementioned works, the approach detailed
below does not aim at devising a decision set in one step and instead
consists of two phases.
The first phase sequentially enumerates all individual minimal rules
given an input dataset.
The second phase computes a minimum-size subset of rules (either in
terms of the number of rules or the total number of literals in use)
that \emph{covers} all the training data instances.
This way the approach trades off large encoding size and thus
potentially hard SAT oracle calls for computing a complete decision
set with a (much) larger number of simpler oracle calls, each
computing an individual rule, followed by solving the set cover
problem.
This algorithmic setup is, in a sense, inspired by the effectiveness
of the standard clausal formula minimization
approach~\cite{quine-amm52,quine-amm55,mccluskey-bstj56,brayton-bk84,espresso-page}.

\subsection{Decision Sets as DNF Formulas} \label{sec:dnf}
First, recall that we consider binary classification, i.e.\
$\fml{C}=\{\ominus,\oplus\}$ but the ideas of this section can be
easily adapted to multi-class problems, e.g.\ by using \emph{one-hot
encoding}~\cite{scikitlearn-full}.
Next, let us split the set of training examples $\fml{E}$ into the
sets of examples \epos and \eneg for the respective classes s.t.
$\fml{E}=\epos \cup \eneg$ and $\epos \cap \eneg = \emptyset$.

Recall that a decision set is an unordered set of \emph{if-then} rules
$\pi\Rightarrow c$, each associating a set of literals $\pi \in
\fml{R}$, $\fml{R}=\prod_{r=1}^{K}\{f_r,\neg f_r,\dcare\}$, over the
feature-values present in rule $\pi$ with the corresponding class $c
\in \{\ominus,\oplus\}$.
Following~\cite{ipnms-ijcar18}, observe that each set of literals
$\pi$ in the rule forms a term and so every class
$c_i\in\{\ominus,\oplus\}$ in a decision set can be represented
logically as a disjunction of terms, each term representing a
conjunction of literals in~$\pi$.

\begin{example} \label{ex:dnf}
  Consider our example dataset shown in Example~\ref{ex:data}.
  Assume that features \emph{Day}, \emph{Venue}, \emph{Weather}, and
  \emph{TV Show} are represented with Boolean variables $f_1$, $f_2$,
  $f_3$, and $f_4$, respectively.
  Observe that all the features $f_r$, $r\in[4]$, in the example are
  binary, and thus each value for feature $f_r$ can be represented
  either as literal $f_r$ or literal $\neg{f_r}$.
  Let us map the original feature values to $\{0,1\}$ such that the
  alphabetically-first value is mapped to $0$ while the other is
  mapped to $1$.
  The classes \emph{No} and \emph{Yes} are mapped to $\ominus$ and
  $\oplus$, respectively.
  As a result, our dataset becomes
  \begin{center}
  \begin{tabular}{ccccc}
  \toprule
  $f_1$ & $f_2$ & $f_3$ & $f_4$ & $c$ \\
  \midrule
  0 & 1 & 1 & 0 & $\ominus$ \\
  1 & 0 & 1 & 0 & $\oplus$ \\
  1 & 0 & 1 & 0 & $\oplus$ \\
  1 & 0 & 0 & 1 & $\ominus$ \\
  \bottomrule
  \end{tabular}
  \end{center}
  Using this binary dataset and the first decision set from
  Example~\ref{ex:data}
  %
  %
  the classes $c=\ominus$ and $c=\oplus$ are represented as the DNF
  formulas
  $\nds \triangleq(f_4) \lor (\neg{f_1})$ and $\pds
  \triangleq(\neg{f_4} \land f_1)$
  \qed
\end{example}

In this work, we follow~\cite{ipnms-ijcar18,yisb-cp20} and compute
minimum-size decision sets in the form of disjunctive representations
\nds and \pds of classes $\ominus$ and $\oplus$.
%
Even though it is simpler to construct rules for one class when there
are only two classes~\cite{meel-cp18,meel-aies19}, 
computing both \nds and \pds achieves better interpretability.
Specifically, if both classes are explicitly represented, it is
relatively easy to extract explicit and succinct explanations for any
class, but this is not the case when only one class is computed.
This approach also immediately extends to problems with three or more
classes.

Without loss of generality, we focus on computing a disjunctive
representation \pds for the class $c=\oplus$.
The same reasoning can be applied to compute \nds.
The target DNF \pds must be consistent with the training data, i.e.\
every term $\pi \in \pds$ must (1)~\emph{agree with} at least one
example $e_i\in\epos$ and (2)~\emph{discriminate} all examples
$e_j\in\eneg$.
Furthermore, each term $\pi\in\pds$ must be \emph{irreducible}, meaning
that any subterm $\pi'\subsetneq\pi$ does not fulfill one of the two
conditions above.

\subsection{Learning Rules} \label{sec:prime}
This section describes the first phase of the proposed approach,
namely, how a term $\pi$ satisfying both of the conditions above can
be obtained separately of the other terms.
Every term is computed as a MaxSAT solution to a partial CNF formula
\begin{equation} \label{eq:drail}
  \psi\triangleq\fml{H}\land\fml{S}
\end{equation}
with $\fml{H}$ and $\fml{S}$ being the
\emph{hard} and \emph{soft} parts, described below.

Consider two sets of Boolean variables $P$ and $N$, $|P|=|N|=K$.
For every feature $f_r$, $r\in[K]$, define variables $p_r\in P$ and
$n_r\in N$.
The idea is inspired by the \emph{dual-rail encoding} (DRE) of propositional
formulas~\cite{bryant-dac87}, e.g.\ studied in the context of logic
minimization~\cite{mfmso-ictai97,jmsss-jelia14}.
Variables $p_r$ and $n_r$ are referred to as \emph{dual-rail
variables}.
We assume that $p_r=1$ iff $f_r=1$ while $n_r=1$ iff $f_r=0$.
Moreover, for every feature $f_r$, $r\in[K]$, a hard clause is added
to $\fml{H}$ to forbid the feature taking two values at once:
\begin{equation} \label{eq:atm1}
  \forall_{r\in[K]}\;\;(\neg{p_r} \lor \neg{n_r})
\end{equation}
The other combinations of values for $p_r$ and $n_r$ encode the fact
that feature $r$ occurs in the target term $\pi$ positively,
negatively, or does not occur at all (when $p_r=n_r=0$).

The set of soft clauses $\fml{S}$ represents a preference to discard
the features from a target term $\pi$ and thus contains a pair of soft
unit clauses expressing that preference:
\begin{align} \label{eq:soft}
  \fml{S}\triangleq\left\{(\neg{p_r}), (\neg{n_r})\;|\; r \in [K]\right\}
\end{align}
By construction of $\fml{S}$ and given a MaxSAT solution for
\eqref{eq:drail}, the target term $\pi$ is composed of all features
$f_r$, for which one of the dual-rail variables (either $p_r$ or
$n_r$) is assigned to 1 by the solution, i.e.\ the corresponding soft
clauses are \emph{falsified}.
%

\paragraph{Discrimination Constraints.}
Every example $e_j=(\mbf{v}_j,\ominus)$ from $\eneg$ must be
discriminated.
This can be enforced by using a clause
$(\bigvee_{r\in[K]}{f_r^{1-v_{jr}}})$, where constant $v_{jr}$ is the
value of the $r^\text{th}$ feature in example~$e_j$.
%
To represent this in the dual-rail formulation, we add the following
hard clauses to~$\fml{H}$:
\begin{align} \label{eq:discr}
  \forall_{j\in[|\eneg|]}\;\;\bigvee_{r\in[K]}{\delta_{jr}},
\end{align}
where $\delta_{jr}$ is to be replaced by dual-rail variable $p_r$ if
$v_{jr}=0$ and replaced by the opposite dual-rail variable $n_r$ if
$v_{jr}=1$.

\begin{example} \label{ex:discr}
  Consider instance $e_1=(f_1=0,f_2=1,f_3=1,f_4=0)\in\eneg$ of the
  running example.
  To discriminate it, we add a hard clause $(p_1 \lor n_2 \lor n_3
  \lor p_4)$.
  Indeed, to satisfy this clause, we have to pick one of the literals
  discriminating example $e_1$, e.g.\ if $p_1=1$ then literal $f_1$
  occurs in term $\pi$, which discriminates instance $e_1$.
  \qed
\end{example}

\paragraph{Coverage Constraints.}
To enforce that every term $\pi$ covers at least one training instance
of \epos, we can use similar reasoning.
Observe that a term $\pi\in\fml{R}$ covers instance
$e_i=(\mbf{v}_i,c_i) \in \epos$ iff none of its literals discriminates
$e_i$, i.e.\ $f_r^{1-v_{jr}}\not\in\pi$ for any $r\in[K]$.
%
%
For each example $e_i=(\mbf{v}_i,c_i) \in \epos$, we introduce an
auxiliary variable $t_i$ defined by:
\begin{equation} \label{eq:aux}
  t_i \leftrightarrow \neg(\bigvee_{r=1}^K{\delta_{ir}}),
\end{equation}
where $\delta_{ir}$ is to be replaced by dual-rail variable $p_r$ if
$v_{ir}=0$ and replaced by the opposite dual-rail variable $n_r$ if
$v_{ir}=1$.
Now, variable $t_i$ is true iff term $\pi$ covers example $e_i$.

\begin{example}
  Consider instance $e_2=(f_1=1,f_2=0,f_3=1,f_4=0)\in\epos$ of the
  running example.
  Introduce variable $t_2 \leftrightarrow \neg (n_1 \lor p_2 \lor n_3
  \lor p_4)$ as shown above.
  %
  %
  If $t_2=1$, the literals in the target term $\pi$ cannot
  discriminate example~$e_2$.
  \qed
\end{example}

Once auxiliary variables $t_i$ are introduced for each example $e_i
\in \epos$, the hard clause
\begin{align} \label{eq:cover}
  \bigvee\nolimits_{i\in [|\epos|]} t_i
\end{align}
can be added $\fml{H}$ to ensure that any term $\pi$ agrees with at
least one of the training data instances.

The overall partial MaxSAT model \eqref{eq:drail} comprises hard
clauses \eqref{eq:atm1}, \eqref{eq:discr}, \eqref{eq:aux},
\eqref{eq:cover} and also soft clauses \eqref{eq:soft}.
The number of variables used in the encoding is $\fml{O}(K+M)$ while
the number of clauses is $\fml{O}(K \times M)$.
Recall that earlier works proposed encoding with $\fml{O}(N \times M
\times K)$ variables and clauses, which in some situations makes it
hard (or infeasible) to prove optimality of large decision sets.

\begin{example} \label{ex:drail}
  Consider our aim at computing rules for class
  $\oplus$ in the running example.
  By applying the DRE, one obtains the formula
  $\psi=\fml{H}\land\fml{S}$ where
  \begin{equation*}
    %
    \fml{H}=\left\{
      \begin{array}{c}
        (\neg{p_1} \lor \neg{n_1}) \land
        (\neg{p_2} \lor \neg{n_2}) \land \\
        (\neg{p_3} \lor \neg{n_3}) \land
        (\neg{p_4} \lor \neg{n_4}) \land \\
        \\
        (p_1 \lor n_2 \lor n_3 \lor p_4) \land \\
        (n_1 \lor p_2 \lor p_3 \lor n_4) \land \\
        \\
        \left[t_2 \leftrightarrow \neg(n_1 \lor p_2 \lor n_3 \lor p_4)\right] \land \\
        \left[t_3 \leftrightarrow \neg(n_1 \lor p_2 \lor n_3 \lor p_4)\right] \land \\
        (t_2 \lor t_3)
      \end{array}
    \right\}
  \end{equation*}
  and
  \begin{equation*}
    \fml{S}=\left\{
      \begin{array}{c}
        (\neg{p_1}) \land (\neg{n_1}) \land (\neg{p_2}) \land (\neg{n_2}) \land \\
        (\neg{p_3}) \land (\neg{n_3}) \land (\neg{p_4}) \land (\neg{n_4})\;\;\;\; \\
      \end{array}
    \right\}
  \end{equation*}
  \qed
\end{example}



Any assignment satisfying the hard clauses $\fml{H}$ of the dual-rail
MaxSAT formula~\eqref{eq:drail} constructed above defines a term $\pi$
that discriminates all examples of \eneg and covers at least one
example of \epos;
soft clauses $\fml{S}$ ensure minimality of terms $\pi$.
More importantly, one can exhaustively enumerate all
solutions of~\eqref{eq:drail} to compute the set of all such terms
(i.e.\ one can use the standard trick of adding a hard clause
\emph{blocking} the previous solution and ask for a new one until no
more solutions can be found).
Let us refer to this set of terms as \tpos.
Finally, we claim that as soon as exhaustive solution enumeration for
formula~\eqref{eq:drail} is finished, the set of terms \tpos covers
every example $e_i \in \epos$.
The rationale is that if sets \epos and \eneg do not overlap then for
any example $e_i \in \epos$ there is a way to cover it by a term $\pi$
s.t.\ all examples of \eneg are discriminated by $\pi$.
This means that, by construction of~\eqref{eq:drail}, for every
variable $t_i$, the hard part $\fml{H}$ of the formula has a
satisfying assignment assigning $t_i=1$.
(Recall that we assume training data to be \emph{perfectly}
classifiable.)

\begin{example} \label{ex:mcs}
  Consider our running example.
  Observe that a valid solution for formula $\psi$ above is
$\{p_1, n_4\}$ from
  which we can extract a term $\pi=(f_1 \land \neg{f_4})$ for the
  target class $c=\oplus$.
  The term $\pi$ is added to \tpos.
  Observe that $\pi$ covers both examples $e_2$ and $e_3$ and
  discriminates examples $e_1$ and $e_4$ from class $\ominus$.
  \qed
\end{example}

\subsection{Smallest Rule Cover} \label{sec:cover}
Once the set \tpos of all terms for class $c=\oplus$ is obtained, the
next step of the approach is to compute a \emph{smallest size} cover
\pds of the training examples \epos.
Concretely, the problem is to select the smallest size subset \pds of
\tpos that covers all the training examples.
The size can be the either the number of terms used or the total
number of literals used in \pds.
Therefore, the problem to solve is essentially the \emph{set cover
problem}~\cite{karp-ccc72}.
Assume that $|\tpos|=L$ and create a Boolean variable $b_j$
for every term $\pi_j \in \tpos$, indicating that rule $j$ is selected.
Also, consider $L \times M'$, $M'=|\epos|$, Boolean constant values
$a_{ij}$ s.t.\ $a_{ij}=1$ iff term $j$ covers example $i$.
Then, the problem of computing the cover with the fewest number of
terms can be stated as:
\begin{align}
  \text{minimize} & \sum_{j=1}^L{b_j} \label{obj:number} \\
  \text{subject to} & \sum_{j=1}^L{a_{ij}\cdot b_j}\geq 1, \forall i
  \in [M'] \label{eq:number_constraint}
\end{align}

Alternatively,
the objective function can be modified
to minimize the total number of literals.
Concretely, create a constant $s_j \in \mbb{Z}$ s.t.\ $s_j=|\pi_j|$,
$\pi \in \tpos$.
The problem is then to
\begin{align}
  \text{minimize} & \sum_{j=1}^L{s_j\cdot b_j} \label{obj:size} \\
  \text{subject to} & \sum_{j=1}^L{a_{ij}\cdot b_j}\geq 1, \forall i
  \in [M'] \label{eq:size_constraint}
\end{align}

\begin{example} \label{ex:cover}
  For our running example, $|\tpos|=4$, with terms being: \[
    \tpos=\left\{
      (f_1 \land f_3), (f_1 \land \neg{f_4}), (\neg{f_2} \land f_3),
      (\neg{f_2}, \neg{f_4}) \\
    \right\} \]
  The set cover problem for $c=\oplus$ can be seen as the
  table:
  \begin{center}
    \small
    \begin{tabular}{ccccccccc}
      \toprule
      & $\pi_1$ & $\pi_2$ & $\pi_3$ & $\pi_4$ \\
      \midrule
      \multirow{2}{*}{$a_{ij}$} & 1 & 1 & 1 & 1 \\
                                & 1 & 1 & 1 & 1 \\
      \midrule $s_j$ & 2 & 2 & 2 & 2 \\ \bottomrule
    \end{tabular}
  \end{center}
  This example has trivial solutions because
  every term covers all examples of \epos, i.e., every $a_{i,j} = 1$
  When minimizing the number of terms using the set cover problem \eqref{obj:number} and \eqref{eq:number_constraint}, every optimal solution contains exactly one term.
  %
  When minimizing the number of literals using the set cover problem \eqref{obj:size} and \eqref{eq:size_constraint}, every optimal solution again
  contains exactly one term, and this term has size 2.

  Now, consider class  $c=\ominus$; $|\tneg|=4$, with terms being: \[
    \tneg=\left\{
      (\neg{f_1}), (f_2), (\neg{f_3}), (f_4)
    \right\}
  \]
  Then the set cover problem for $c=\ominus$ can be seen as the
  following table:
  \begin{center}
    \small
    \begin{tabular}{ccccccccc}
      \toprule
      & $\pi_1$ & $\pi_2$ & $\pi_3$ & $\pi_4$ \\
      \midrule
      \multirow{2}{*}{$a_{ij}$} & 1 & 1 & 0 & 0 \\
                                & 0 & 0 & 1 & 1 \\
      \midrule $s_j$ & 1 & 1 & 1 & 1 \\ \bottomrule
    \end{tabular}
  \end{center}
  In our case, one valid solution picks columns $\pi_1$ and
  $\pi_3$ as both of them together cover \eneg.
  Thus, when minimizing the number of terms, the fewest number of columns
  to pick, such that every row has at least one value, is 2, i.e.,
  any optimal solution has cost 2
  ($1+1$).
  \qed
\end{example}

\subsubsection{Breaking Symmetric Rules.} \label{sec:bsymm}
Observe in Example~\ref{ex:cover} that the two terms
$\pi_1$ and $\pi_2$ in \tpos cover the same examples from \epos.
In the context of the set cover problem, such terms are described as
\emph{symmetric}.
It is straightforward that at most one term in a set of symmetric terms can appear in a solution
because of the minimization in the set cover problem.

Symmetric terms can become an issue if the total number of terms is
\emph{exponential} on the number of features.
This kind of repetition can be avoided by using the instance coverage
variables $t_i$.
Concretely, given a term $\pi \in \tpos$ covering a set $\epos'
\subset \epos$ of data instances, one can add a clause $(\bigvee_{i
\in \epos \setminus \epos'}{t_i})$ enforcing that any terms discovered
later must cover at least one instance $e_i$ uncovered by term~$\pi$.

While the terms are symmetric for objective~\eqref{obj:number}, there
is a dominance relation for objective~\eqref{obj:size}.
Consider a term $\pi$ with the same coverage as another term $\rho$
s.t.\ $|\rho|\geq|\pi|$ --- term $\pi$ \emph{dominates} term $\rho$.
Given a set cover solution $S\cup \{\rho\}$, we can always replace
$\rho$ by $\pi$ to get a \emph{no worse} solution $S\cup \{\pi\}$.
Therefore, term $\rho$ can be ignored during selection.

Because term enumeration is done with MaxSAT, i.e.\ smaller terms come
first, we can use the same method above for symmetry to eliminate
dominated terms, since a dominating term (a smallest term with the
same coverage) will always be discovered first.
Clearly, optimality for objectives~\eqref{obj:number} and
\eqref{obj:size} is still guaranteed if breaking symmetric rules is
applied.

\begin{example} \label{ex:bsymm}
  By breaking symmetric terms, the enumeration procedure computes only
  one term for class $\oplus$ and two terms for class $\ominus$.
  (Recall that we previously got $|\tpos|=|\tneg|=4$.)
  Note that all of them are included in the solutions for the
  corresponding set cover problems.
  \qed
\end{example}

\begin{figure*}[!t]
  \begin{subfigure}[b]{0.6\textwidth}
    \centering
    \includegraphics[width=0.94\textwidth]{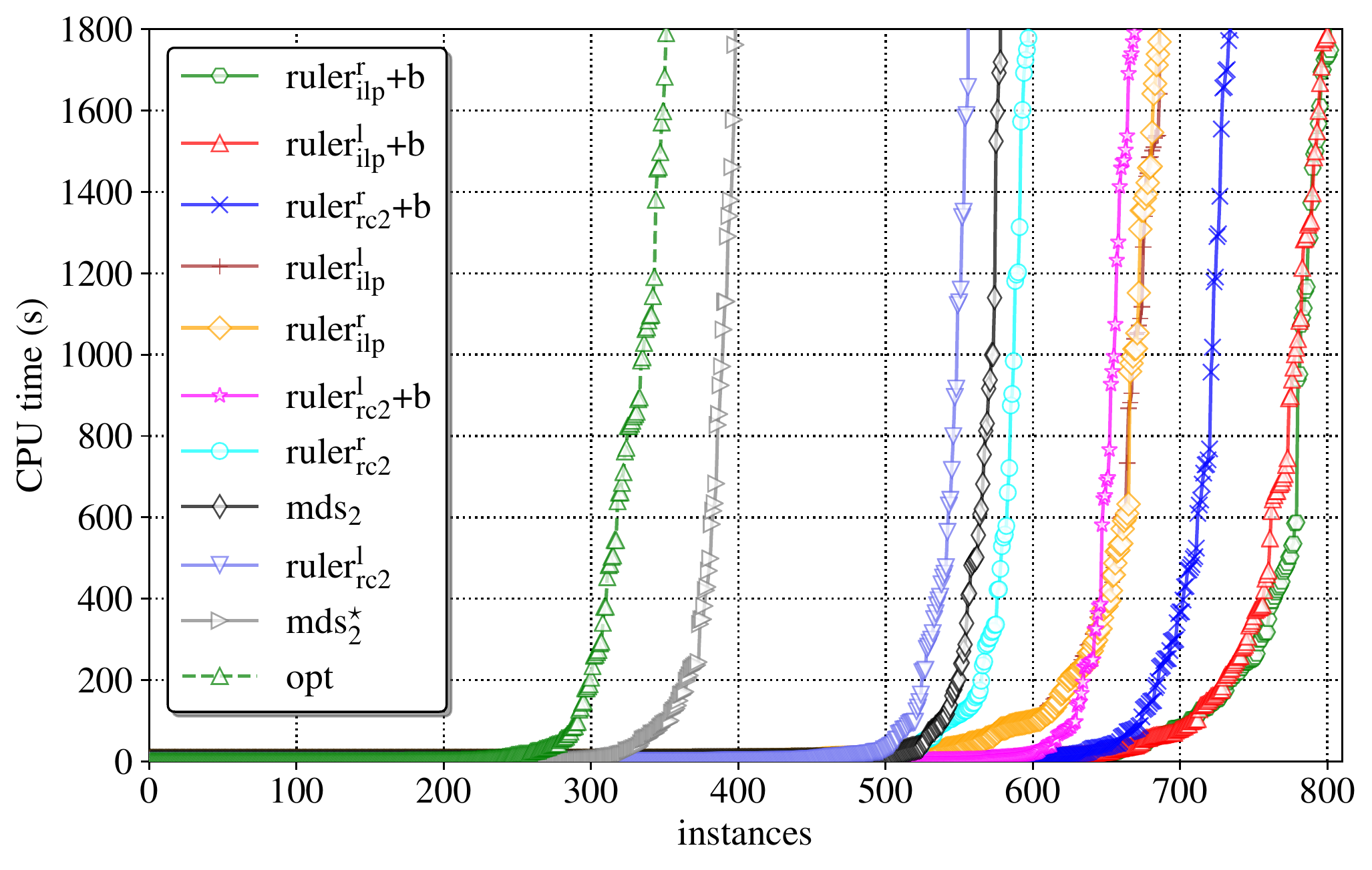}
    \caption{Raw performance}
    \label{fig:cact}
  \end{subfigure}%
  %
  %
  \begin{subfigure}[b]{0.4\textwidth}
    \centering
    \includegraphics[width=0.92\textwidth]{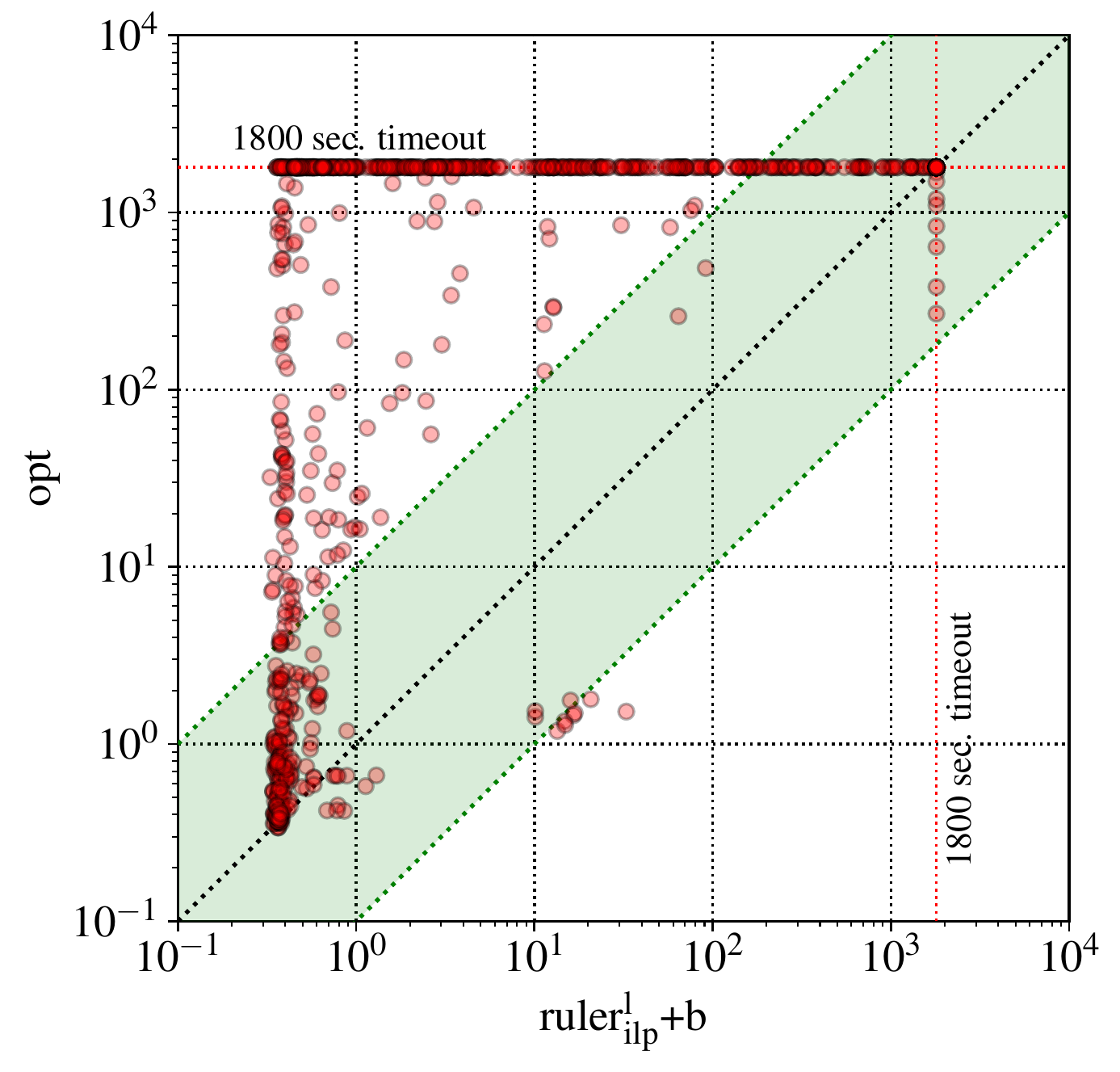}
    \caption{Detailed runtime comparison of \mpr{ilp}{l} vs.~\emph{opt}}
    \label{fig:det}
  \end{subfigure}
  \caption{Scalability of the competitors.}
  \label{fig:perf}
\end{figure*}

\section{Experimental Results} \label{sec:res}
%
%
This section evaluates the proposed rule enumeration based approach in
terms of scalability and compares it with the state of the art of
SAT-based learning of minimum-size decision sets on a variety of
publicly available datasets.
The experiments were performed in Debian Linux on an Intel Xeon
Silver-4110~2.10GHz processor with 64GByte of memory.
Following the setup of recent work~\cite{yisb-cp20}, the time limit
was set to 1800s for each individual process to run.
The memory limit was set to 8GByte per process.

\subsubsection*{Prototype Implementation and Selected Competition.}
A prototype\footnote{Available as part of
\url{https://github.com/alexeyignatiev/minds}.} of our rule
enumeration based approach was developed as a set of Python scripts,
in the following referred to as \emph{ruler}.
The implementation of rule enumeration was done with the use of the
state-of-the-art MaxSAT solver RC2~\cite{imms-sat18,imms-jsat19},
which proved to be the most effective in MaxSAT model
enumeration~\cite{mse20}.
As a result, the terms are computed in a sorted fashion, i.e.\ the
smallest ones come first.
For the second phase of the approach, i.e.\ computing the set
cover, we attempted to solve the problem both (1)~with the RC2 MaxSAT
solver and (2)~with the Gurobi ILP solver~\cite{gurobi}.
The corresponding configurations of the prototype are called
\mpr{rc2}{\ast} and \mpr{ilp}{\ast}, where `$\ast$' can either be
\emph{`r'} or \emph{`l'} meaning that the solver minimizes either the
number of rules or the total number of literals.\footnote{We also
tried using Gurobi for the first phase but it was significantly
outperformed by the MaxSAT-based solution.}
Configurations \mprb{\ast}{\ast} apply symmetry breaking constraints
to reduce the number of implicant rules computed in the first phase of
the approach.
Finally, all configurations were set to compute explicit optimal DNF
representations for all classes given a dataset.

The competiting approaches include SAT-based
methods~\cite{ipnms-ijcar18} $MinDS_2$ and $MinDS_2^\star$ referred
to as $mds_2$ and $mds_2^\star$.
The former tool computes the fewest number of rules while the latter
lexicographically minimizes the number of rules and then the number of
literals.
The second competitor is a recent SAT-based approach that minimizes the
total number of literals in the model~\cite{yisb-cp20}, in the
following referred to as \emph{opt}.
Note that as the main objective of this experimental assessment is to
demonstrate \emph{scalability} of the proposed approach, the methods
for computing sparse decision
sets~\cite{meel-cp18,meel-aies19,yisb-cp20} are intentionally
excluded due to the significant difference in the
problem they tackle.
%

\subsubsection{Benchmarks.}
All datasets considered in the evaluation were adopted from
~\cite{yisb-cp20} and used unchanged. These datasets originated from
the UCI Machine Learning Repository~\cite{uci} and the Penn Machine
Learning Benchmarks~\cite{pennml}.
The total number of datasets is 1065.
The number of \emph{one-hot encoded}~\cite{scikitlearn-full} features
(training instances, resp.) per dataset in the benchmark suite varies
from 3 to 384 (from 14 to 67557, resp.).
Also, since \emph{ruler} can handle only perfectly classifiable data,
it processes each training dataset by keeping the largest consistent
(non-overlapping) set of examples.
This technique is applied in~\cite{ipnms-ijcar18,yisb-cp20} as well,
which enables one to achieve the highest possible accuracy on the
training data.
Motivated by one of the conclusions of~\cite{yisb-cp20} stating that
perfectly accurate decision sets, if successfully computed, are
significantly more accurate than sparse and also heuristic models,
here we do not compare test accuracy of the competitors -- we assume
test accuracy to be (close to) identical for all the considered
approaches.

\begin{figure*}[!t]
  \begin{subfigure}[b]{0.5\textwidth}
    \centering
    \includegraphics[width=0.63\textwidth]{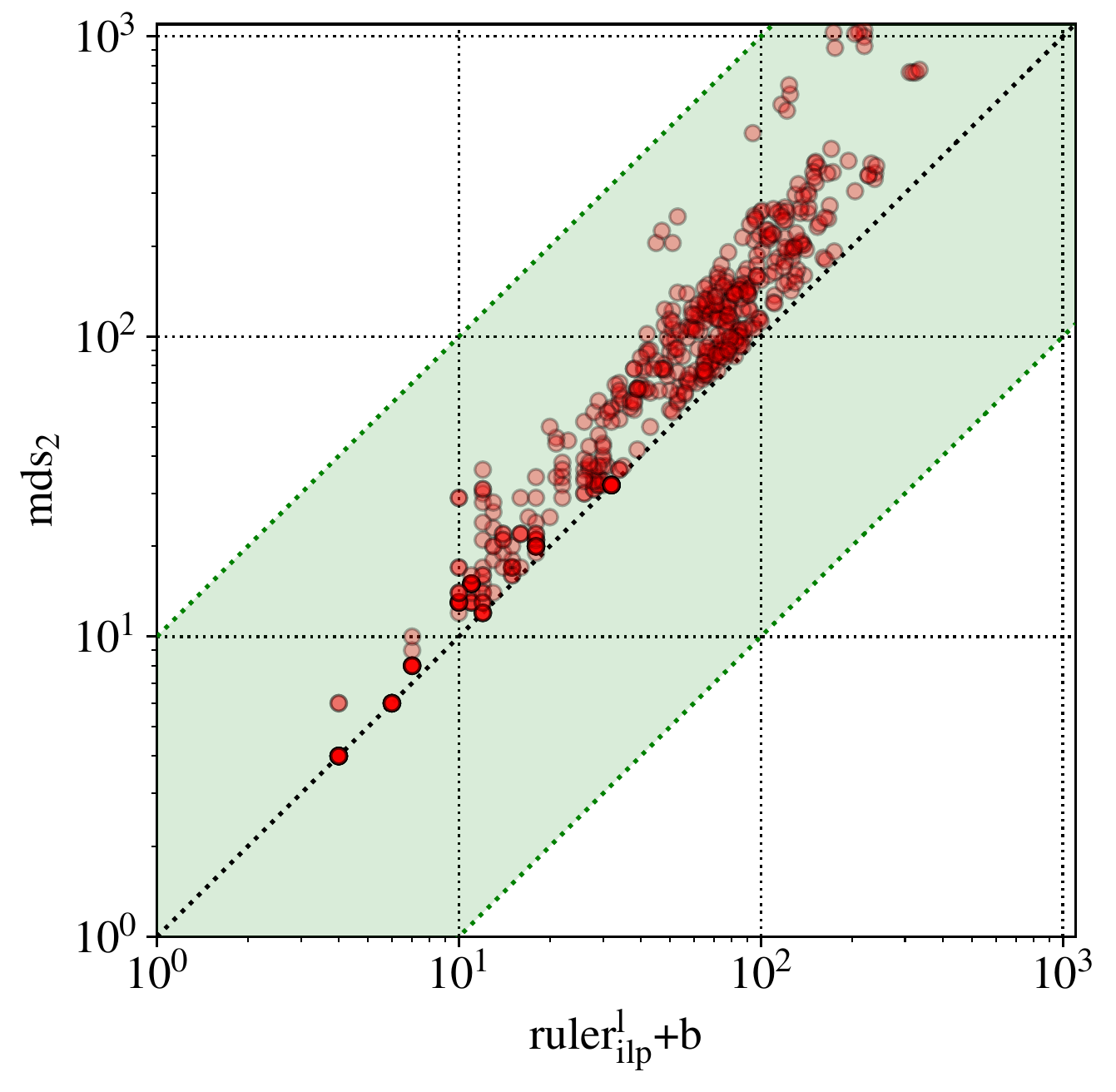}
    \caption{Literals or rules: \mpr{ilp}{l} vs.~$mds_2$}
    \label{fig:size1}
  \end{subfigure}%
  %
  %
  \begin{subfigure}[b]{0.5\textwidth}
    \centering
    \includegraphics[width=0.63\textwidth]{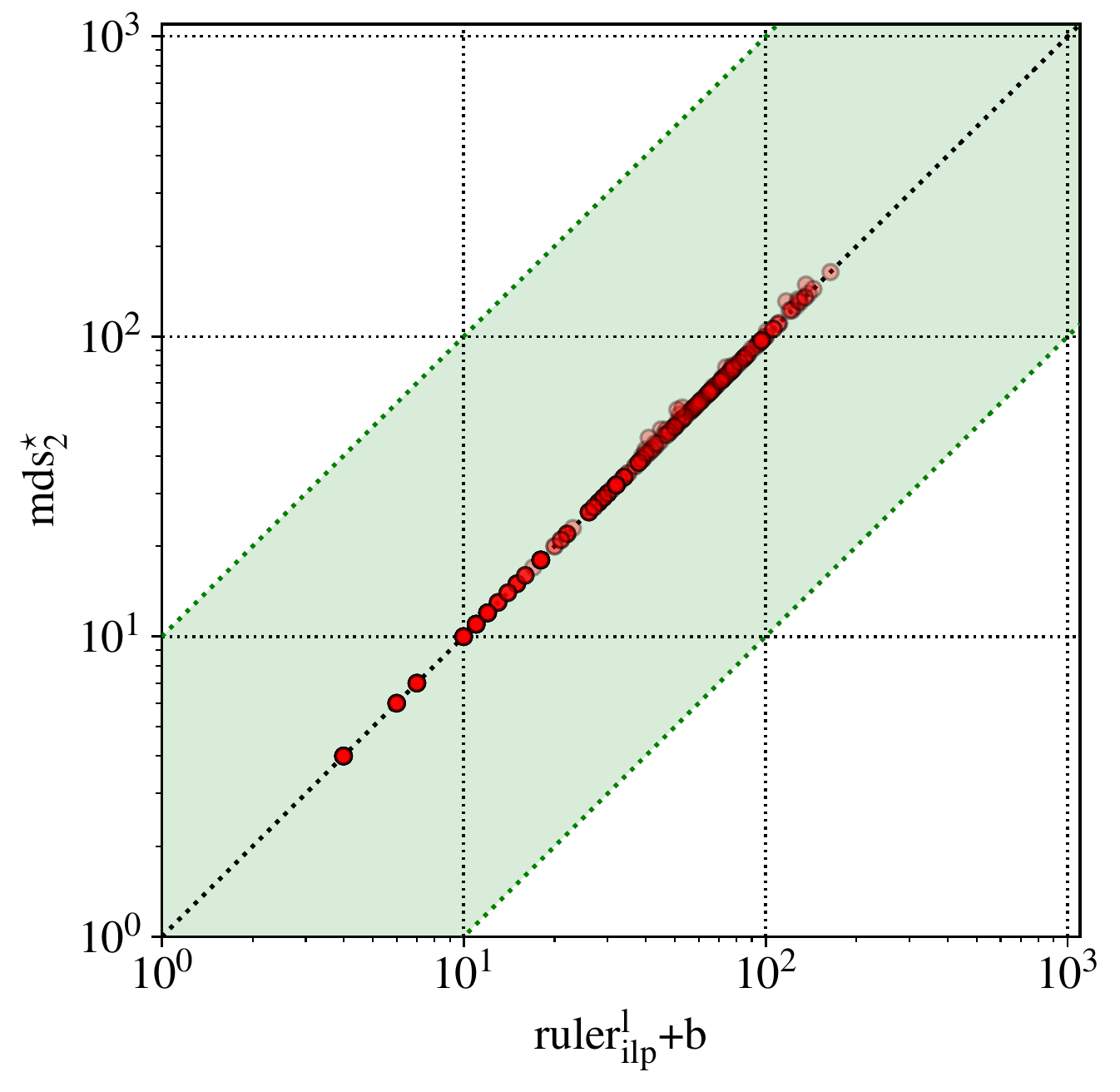}
    \caption{Literals or lexicographic: \mpr{ilp}{l} vs.~$mds_2^\star$}
    \label{fig:size2}
  \end{subfigure}
  \caption{Model size comparison.}
  \label{fig:size}
\end{figure*}

\subsubsection{Raw Performance.}
Figure~\ref{fig:cact} shows scalability of all the selected
approaches.
Observe that the mixed solution \mprb{ilp}{r} demonstrates the best
performance being able to train decision sets for 802 datasets.
Second best approach is \mprb{ilp}{l} and copes with 800 benchmarks.
Pure MaxSAT-based \mprb{rc2}{r} and \mprb{rc2}{l} perform worse with
734 and 669 instances solved.
This is not surprising because the structure of set cover problems
naturally fits the capabilities of modern ILP solvers.

Disabling symmetry breaking constraints affects the performance of all
configurations of \mpr{\ast}{\ast}, which drops significantly.
When it is disabled, the best such configuration (\mpr{ilp}{l}) solves
686 benchmarks while the worst one (\mpr{rc2}{l}) tackles 556.
Here, we should say that the \emph{maximum} and \emph{average} number
of rules enumerated if symmetry breaking is disabled is 326399 and
19604.4, respectively.
Breaking symmetric rules decreases these numbers to 8865 and 563.7,
respectively.

As for the rivals of the proposed approach, the best of them ($mds_2$)
is far behind \mprb{ilp}{\ast} and successfully trains 578 models even
though it targets rule minimization, which is arguably a much simpler
problem.
Another competitor ($mds_2^\star$) lexicographically minimizes the
number of rules and then the number of literals, which is
unsurprisingly harder to deal with, as it solves 398 benchmarks.
Finally, the worst performance is demonstrated by \emph{opt}, which
learns 351 models.
We reemphasize that both \emph{opt} and \mprb{ilp}{l} compute minimum-size
decision sets in terms of the number of literals.
However, the proposed solution outperforms the competition by 449
benchmarks.
Note that due to the large encoding size, \emph{opt} ($mds_2$, resp.)
is practically limited to optimal models having a few dozens of
literals (rules, resp.).
There is no such limitation in \mpr{\ast}{\ast} -- in our experiments,
it could obtain minimum-size models having thousands of literals in
total within the given time limit.
The runtime comparison for \mprb{ilp}{l} and \emph{opt} is detailed in
Figure~\ref{fig:det} -- observe that except for a few outliers,
\mprb{ilp}{l} outperforms the rival by up to four orders of magnitude.

\subsubsection{Rules vs.~Literals.}
Here we demonstrate that literal minimization results in smaller and
thus more interpretable models than rule minimization.
Figure~\ref{fig:size1} compares the total number of literals in the
models of $mds_2$ and \mprb{ilp}{l} while Figure~\ref{fig:size2}
compares the model sizes for $mds_2^\star$ and \mprb{ilp}{l}.
To make a valid comparison, we used only instances solved by both
approaches in each pair.
Among the datasets used in of Figure~\ref{fig:size1}, the average
number of literals obtained by $mds_2$ and \mprb{ilp}{l} is 116.2 and
62.2, respectively -- the advantage of literal minimization is clear.
Also, as shown in Figure~\ref{fig:size2}, lexicographic optimization
results in models almost identical in size to the models produced by
\mprb{ilp}{l}.
This suggests that applying the approach of $mds_2^\star$ may in
general pay off in terms of solution size, by significantly
sacrificing scalability compared to $mds_2$ and, more importantly, to
\mprb{ilp}{l} (398 vs.~578 vs.~800 instances solved, which represents
37.4\%, 54.7\%, and 75.1\% of all 1065 benchmarks, respectively).

\section{Conclusions} \label{sec:conc}
This paper has introduced a novel approach to learning minimum-size
decision sets based on individual rule enumeration.
The proposed approach has been motivated by the standard twofold
methods applied in two-level logic
minimization~\cite{quine-amm52,quine-amm55,mccluskey-bstj56} and split
the problem into two parts: (1)~exhaustively enumerating individual
rules followed by (2)~solving the set cover problem.
The basic approach has been additionally augmented with symmetry
breaking, enabling us to significantly reduce the number of rules
produced.
The approach has been applied to computing minimum-size decision sets
both in terms of the number of rules and in terms of the total number
of literals.
The proposed approach has been shown to outperform the state of the
art in logic-based learning of minimum-size decision sets by a few orders
of magnitude.

As the proposed approach targets computing \emph{perfectly accurate}
decision sets, a natural line of future work is to examine ways of
applying it to computing \emph{sparse} decision sets that trade off
accuracy for size.
Another line of work is to address the issue of potential rule overlap
wrt.~the proposed approach.
Finally, it is of interest to apply similar rule enumeration
techniques for devising other kinds of rule-based ML models, e.g.\
decision lists and decision trees.



\bibliography{refs}

\end{document}